\documentclass{article}
\usepackage{spconf,amsmath,graphicx}
\usepackage{makecell}
\usepackage{multirow, stfloats}
\usepackage{color,soul}
\usepackage{hyperref}
\usepackage{multirow, stfloats}
\usepackage{booktabs}
\usepackage{amssymb}
\usepackage{bbding}

\title{Assessing Phrase Break of ESL speech with Pre-trained Language Models}
\name{Zhiyi Wang*, Shaoguang Mao*, Wenshan Wu, Yan Xia\thanks{*Equal Contribution.}}
\address{Microsoft Research Asia}
\begin{document}
%
\maketitle
\begin{abstract}
This work introduces an approach to assessing phrase break in ESL learners' speech with pre-trained language models (PLMs). Different with traditional methods, this proposal converts speech to token sequences, and then leverages the power of PLMs. There are two sub-tasks: overall assessment of phrase break for a speech clip; fine-grained assessment of every possible phrase break position. Speech input is first force-aligned with texts, then pre-processed to a token sequence, including words and associated phrase break information. The token sequence is then fed into the pre-training and fine-tuning pipeline. In pre-training, a replaced break token detection module is trained with token data where each token has a certain percentage chance to be randomly replaced. In fine-tuning, overall and fine-grained scoring are optimized with text classification and sequence labeling pipeline, respectively. With the introduction of PLMs, the dependence on labeled training data has been greatly reduced, and performance has improved.
\end{abstract}
\begin{keywords}
Phrase break, computer-aided language learning, ESL speech, Pre-trained language models
\end{keywords}
\section{Introduction}
\label{sec:intro}
Proper phrase break is crucial to oral performance\cite{fach1999comparison} and is always a challenge for English as a Second Language (ESL) learners. There has been considerable research in computer-aided language learning (CALL)\cite{mao2019nn, lin2021improving, mao2022universal, hu2015improved}. As for the phrase break assessment, there are two main categories: 1) break feature extraction and modeling\cite{fu2022using, sabu2018automatic}. 
2) modeling against reference speech\cite{xiao2017proficiency, proencca2019teaching}.
For example, a method was proposed to evaluate break by computing similarity between the assessed speech with utterances from native speakers or Text-to-Speech (TTS) system \cite{xiao2017proficiency}.

\begin{figure}[t]
\centering
\includegraphics[scale=0.5]{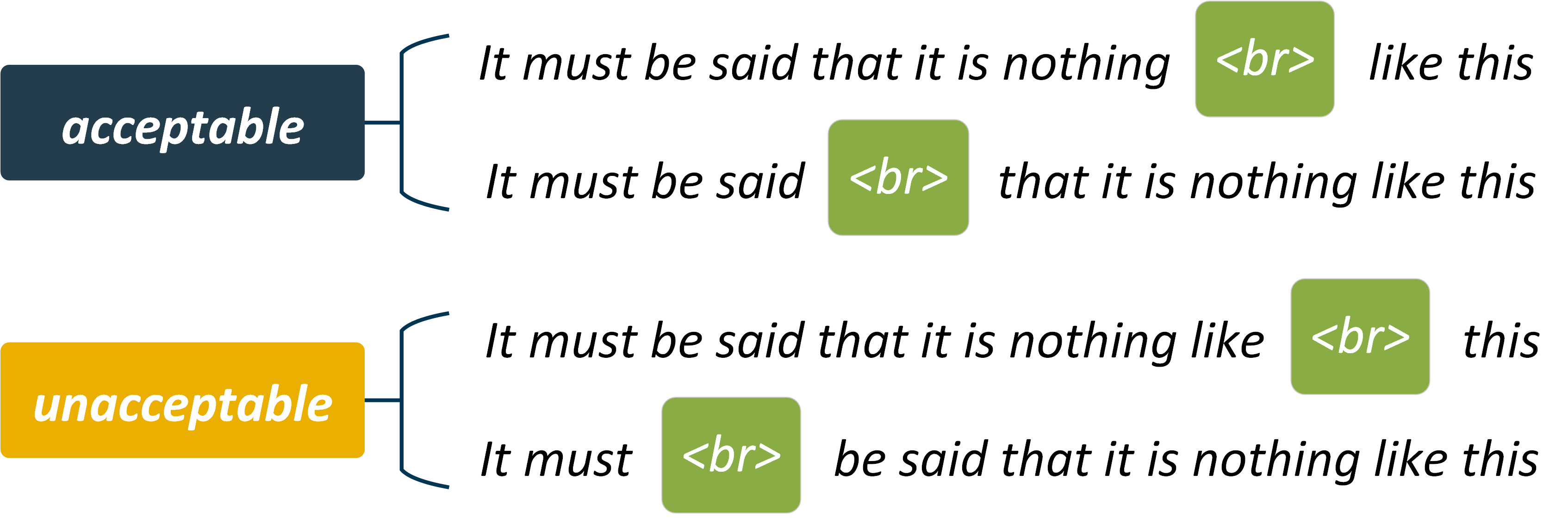}
\caption{Example of diverse phrase break patterns(${<}$br${>}$ represents a phrase break)}\label{fig:break patterns}
\end{figure}

Although modeling against reference speech\cite{xiao2017proficiency, proencca2019teaching} is an effective approach to assessing speech performance,  some errors unavoidably occur when it comes to handle diverse phrase break cases. As shown in Figure~\ref{fig:break patterns}, the correct phrase break patterns for the same text are various, and thus it is not to say that the phrasing is incorrect if it is different with template audios. The previous works fail to consider this fact. Instead, they model the phrase break like a fixed pattern prediction.

Meanwhile, a large scale of high-quality human-labeled data is required for traditional methods. However, the subjective labeling is costly and the labeling consistency is hard to satisfy\cite{mao2019nn, zhang2021speechocean762, meng2010development}. How to construct robust models with small dataset is still under research.

Phrase break prediction is a traditional task in the TTS area\cite{futamata2021phrase, kunevsova2022detection, liu2020exploiting, rendel2016using}. In Futamata's work\cite{futamata2021phrase}, 
a phrase break prediction method is proposed that combines implicit features extracted from BERT and explicit features extracted from
BiLSTM with linguistic features. Although the goals of phrase break prediction and phrase break assessment are different, the second one being much harder considering the diverse break facts, we can refer to the idea that the break information can be inferred from input text, and leverage the power of rising pre-trained language models (PLMs)\cite{devlin2018bert, liu2019roberta, dong2019unified}.


This paper presents a new approach to assessing phrase break with PLMs. In particular, there are two sub-tasks: assessment on phrase break for a speech and fine-grained assessment on each break position.

\begin{figure*}[ht]
\centering
\includegraphics[scale=4]{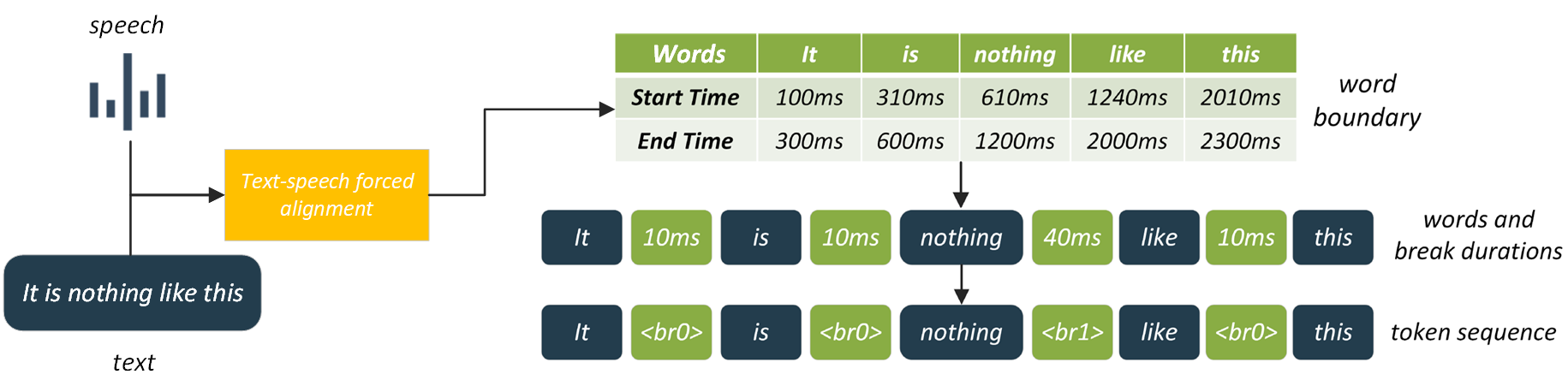}
\caption{Overview of the pre-processing. The speech-text forced alignment tool recognizes word boundaries and the duration between adjacent words. Then, the final token sequence is obtained by converting the duration to break tokens with Table~\ref{tab:break_type} .}\label{fig:tokenization}
\end{figure*}

First, each speech is processed into a token sequence with text-speech forced alignment\cite{mathad2021impact, moreno1998recursive, moreno2009factor}, referencing Figure~\ref{fig:tokenization}. The token sequence consists of words and associated phrase break tokens (break duration information for each between-words interval). Then, in the pre-training stage, a replaced break token detection strategy is employed. Each break token from the original sample has 15\% chance to be replaced by other break tokens. Then, a discriminator is trained with augmented data riding on BERT\cite{devlin2018bert} to identify whether the token sequence is edited. Finally, in the fine-tuning stage, the overall assessment and fine-grained assessment are fine-tuned with text classification and token classification, respectively.

The main contributions are: First, A new pre-training fine-tuning pipeline is proposed for assessing phrase break, where self-supervised learning and the power of PLMs are leveraged, and consequently a robust model is trained with a small dataset. Second, this work takes diverse phrasing patterns into consideration to construct a more precise assessment.

\section{Approach}
\label{sec:format}
\subsection{Task definition}
We use two tasks to demonstrate how PLMs can be applied into assessing phrase break.

One is predicting a rank $r$ for a test speech to indicate its overall performance on phrase break . The other one is that given a speech $S$, consisting $n$ words, predict a rank $r_i$ for each interval $b_i$ between two words on whether the phrase break is appropriate, including whether an existing break is appropriate and whether an expected break is missed.


\subsection{Pre-processing}

To leverage the power of PLMs, the speech clips are first converted to a token sequence with speech-text forced alignment. 

As shown in Figure~\ref{fig:tokenization}, speech-text forced alignment is used to recognize the phrase break and duration between every pair of adjacent words $w_i$ and $w_{i+1}$. Based upon the statistical information and linguists' suggestions, the phrase breaks are categorized into four types, as shown in Table~\ref{tab:break_type}. A speech utterance is then tokenized into a token sequence $T:\left\{w_0,b_0,w_1,...,w_i,b_i,w_{i+1},...,w_n\right\}$, including words and phrase break tokens. 

\subsection{Replaced Break Token Detection}
\label{sec:pre-training}
\begin{figure*}[ht]
\centering
\includegraphics[scale=0.6]{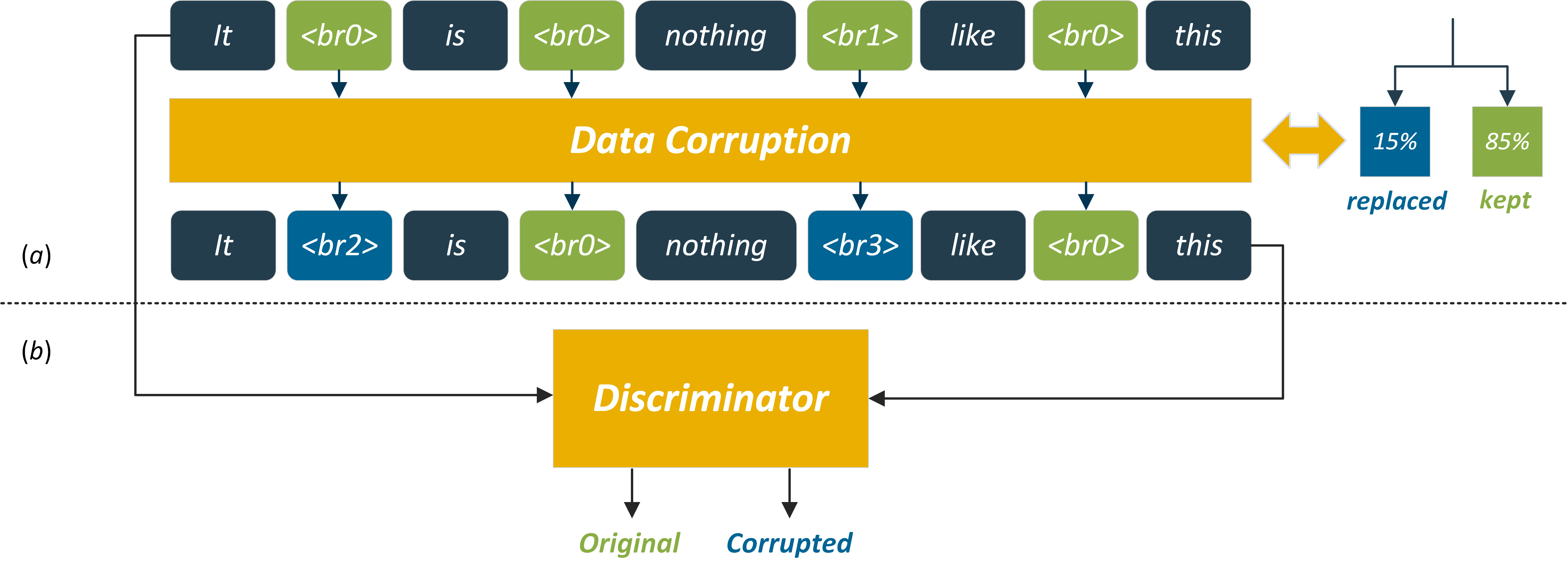}
\caption{An overview of the replaced break token detection pre-training process. Part (a) describes an example of data corruption where each break token in the original token sequence has 15\% chance to be replaced with other tokens. Part (b) is the pre-training stage where a discriminator is trained to distinguish original or corrupted data.}\label{fig:data_corruption}
\end{figure*}

We introduce a pre-training approach named replaced break token detection. As shown in Figure~\ref{fig:data_corruption}, speech recordings by native speakers from TTS corpus are collected as original samples because TTS recordings have good performance in phrase break. Then, each sample is randomly corrupted several times with the strategy that each break token has 15\% chance to be replaced with other kinds of break tokens. 

\begin{table}
\centering
\begin{tabular}{lll}
\hline
\textbf{Type} & \textbf{Duration} & \textbf{Comment}\\
\hline
{br0} & {(0, 10ms]} & {No break} \\
{br1} & {(10ms, 50ms]} & {Slight / Optional break} \\
{br2} & {(50ms, 200ms]} & {Break} \\
{br3} & {(200ms, $+ \infty$)} & {Long break} \\
\hline
\end{tabular}
\caption{The definition of break tokens}
\label{tab:break_type}
\end{table}

With corruption, the real data from TTS recordings represents acceptable break patterns, while corrupted data represents potential errors. Although acceptable phrase breaks are various, we assume that after a series of random replacements on tokens, it is very likely that the original samples have been corrupted to incorrect break patterns.

The pre-training is from BERT as it is trained on a large scale of texts and learns contextual relations between words (or sub-words)\cite{devlin2018bert}. 
A discriminator is trained with cross-entropy loss on the augmented data to predict whether the input sequence has been corrupted. The trained model is called Break-BERT for convenience.

\subsection{Downstream Tasks}
\subsubsection{Overall Assessment}
The overall assessment task is treated as a sequence classification task.
The model predicts a rank $r$ for a token sequence, $r \in R$. 
The model consists of the head of the pre-trained model and a classifier on top, and is trained with cross-entropy loss.

\subsubsection{Fine-grained Assessment}
The fine-grained assessment is treated as a sequence labeling task. An $r_i \in R$ is expected to be assigned to $b_i$. There is a token classification head on top of the hidden-states output from pre-trained model. It is also trained with a cross-entropy loss function. 

\section{Experiments}
\label{sec:pagestyle}
\subsection{Pre-training}
A TTS corpus, LJ Speech\cite{ljspeech17}, is employed to conduct pre-training as the TTS recordings are commonly in good phrase breaks and contain diverse phrase break patterns. The detailed information can be found in Table~\ref{tab:pretrain_dataset}. 

In data corruption, the ratio between the corrupted samples and the original samples is 3:1, i.e. for each original sample, three random corrupted samples are augmented. The pre-training begins from BERT$_\textbf{BASE}$, and a simple linear classifier is added on the top. It is trained with a batch size of 64 for 3 epochs over the dataset. The maximum sequence length is set to 128. We used back propagation and Adam optimizer with a learning rate of 1e-4. After the pre-training, the binary classification results of Break-BERT are reported in Table~\ref{tab:pre-train_experiments}.

\begin{table}[t]\centering
\begin{tabular}{c|c|ccc}
\toprule
\textbf{Dataset}& \textbf{Part} & \textbf{Clips} & \textbf{Words}
&\textbf{Dura.}\\
\midrule
\multirow{2}{*}{LJ Speech} & {Train} & {11.5K} & {192K} & {22.5h}\\
\cline{2-5}
 & {Test} & {500} & {8K} & {1h}\\
\bottomrule
\end{tabular}
\caption{Statistics of pre-training dataset. (Dura. indicates Duration)}
\label{tab:pretrain_dataset}
\end{table}

\begin{table}
\centering
\begin{tabular}{c|cc}
\hline
\textbf{Model} &\textbf{Acc.}  & \textbf{F-score}\\
\hline
{Break-BERT} & {83.9\%} & {89.7\%}\\
\hline
\end{tabular}
\caption{Performance of pre-trained models. (Acc. indicates Accuracy)}
\label{tab:pre-train_experiments}
\end{table}




\begin{table}[h]\centering
\begin{tabular}{c|ccc|c}
\toprule
\textbf{Dataset} &\textbf{Poor} & \textbf{Fair} & \textbf{Great} & \textbf{Total}\\
\hline
{Overall} & {21} & {136} & {643} & {800}\\
\hline
{Fine-grained} & {129} & {644} & {10797} & {11570}\\
\bottomrule
\end{tabular}
\caption{Statistics of downstream datasets.}
\label{tab:finetune_dataset}
\end{table}

\begin{table*}[h]\centering
\begin{tabular}{c|c|ccc}
\toprule
\multicolumn{2}{c|}{\multirow{2}*{\textbf{Assessment Model}}}
& \multicolumn{3}{c}{\textbf{Metric avg.(std)}}\\
\cmidrule(r){3-5}
\multicolumn{2}{c|}{~} &{Acc.} & {F-Score(weighted)}& {F-Score(macro)}\\
\hline 
\multirow{4}*{\textbf{Overall}}&{Bi-LSTM} & {80.2(0.64)} & {76.4(0.96)}& {39.2(0.81)}\\
&{Against-TTS} & {54.4(0.99)} & {61.1(0.71)}& {36.3(0.56)} \\
&{\#BERT} & {80.4(0.65)} & {77.9(0.70)}& {40.9(0.71)} \\
&{\#Break-BERT} & \textbf{82.5(0.50)} & \textbf{81.7(0.57)}& \textbf{52.3(1.05)} \\
\hline
\multirow{4}*{\textbf{Fine-grained}}&{Bi-LSTM} & {92.5(0.39)} & {90.1(0.56)} & {39.9(0.37)}\\
&{Against-TTS} & {70.9(0.26)} & {78.6(0.40)}& {31.1(0.15)} \\
&{\#BERT} & {91.8(0.41)} & {89.0(0.58)} & {39.5(0.41)}\\
&{\#Break-BERT} & \textbf{92.8(0.31)} & \textbf{91.6(0.40)} & \textbf{44.3(0.25)}\\
\bottomrule
\end{tabular}
\caption{Performance of overall and fine-grained assessment models. '\#' stands for 'Fine-tune'.}
\label{tab:overall_scoring}
\end{table*}

\subsection{Downstream Task Corpora}
We collected 800 audio samples from different Chinese ESL learners. Then, two linguists are invited to assess them with overall performance on phrase break and each individual phrase break ranging from 1 (Poor), 2 (Fair), and 3 (Great). If two experts' opinions are inconsistent, an extra linguist will intervene and do the final scoring. The statistics of collected corpus are listed in Table~\ref{tab:finetune_dataset}. All data is available for research.

\begin{figure*}[ht]
\centering
\includegraphics[scale=0.4]{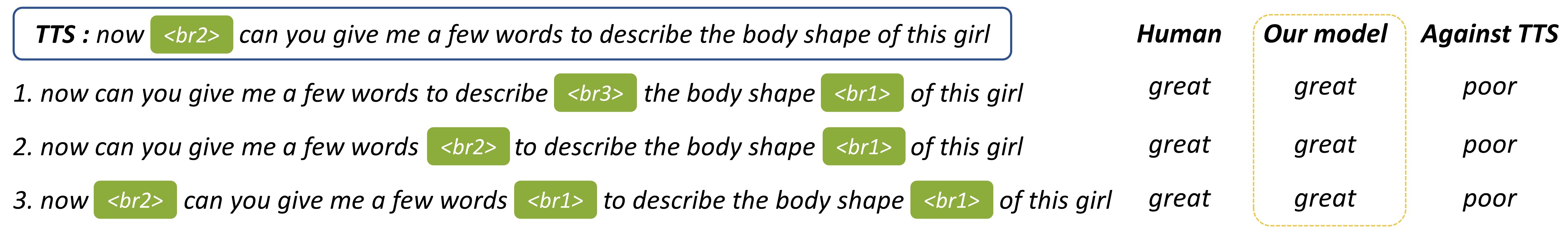}
\caption{The comparison of different approaches dealing with multiple phrase break patterns of the same text. There are some phrase break patterns that differ from TTS output audio but are treated as correct by humans.}\label{fig:break_case_study}
\end{figure*}

\subsection{Experimental Setup}
\textbf{Baselines}
Bi-LSTM+Linear Layer and Bi-LSTM+CRF (Conditional Random Field)\cite{rei2017semi} are set as baselines for overall and fine-grained assessment, separately. We apply Bi-LSTM as a backbone considering it still works well in a relatively small dataset. The hidden layer size is set to 1024. Meanwhile, a direct fine-tuning with downstream data on BERT is conducted to verify the validity of the proposed pre-training process. The baseline models take the identical token sequences by the BERT tokenizer.

Additionally, we adopt the Against-TTS method \cite{xiao2017proficiency} as a baseline and tag the output break similarity score [0, 0.3), [0.3, 0.7), [0.7, 1.0] as poor, fair, great, separately. The adopted TTS system is from Microsoft Cognitive Service en-US-AriaNeural voice.

\textbf{Cross-validation} 
We apply five-fold cross-validation to avoid instability of sampling and report the mean and standard deviation of experiments.


\subsection{Results}

Accuracy, weighted f-score and macro f-score are taken as metrics\cite{zhang2021speechocean762}. As shown in Table~\ref{tab:overall_scoring},  compared with Bi-LSTM, Against-TTS and fine-tuning on BERT, the proposed pre-training fine-tuning greatly improves all metrics.
It is worth mentioning that the Against-TTS system performs much worse than the proposed approach. More discussions are included in the next section.

\section{Discussions}
\subsection{Influence of Pre-training}
The pre-training process takes TTS human recordings as correct samples, where multiple phrase break patterns exist. After a series of random corruptions, the augmented samples are likely to be incorrect in phrasing. After the pre-training on original and constructed incorrect patterns, the discriminator has learned general linguistic patterns and phrase break information through self-supervised learning.

The experiments verified the assumptions. The proposed model yields better results. The knowledge learned from pre-training benefits downstream tasks.


\subsection{How Diverse Breaks are Handled}
The experimental results verified Against-TTS approach's limits on handling multiple possible phrase breaks.  As shown in Table~\ref{tab:confusing_matric}, there are sharp drops of the recall of category 3 (Great) and the precision of category 1 (Poor), while the precision of category 3 (Great) and the recall of category 1 (Poor) are kept. For a test speech, if it shows a different phrase break pattern with reference audio, it tends to be classified as poor even if it is correct. 
\begin{table}[h]\centering
\begin{tabular}{c|c|c|c|c}
\toprule
~ & \multicolumn{2}{c|}{\textbf{Against-TTS}} & \multicolumn{2}{c}{\textbf{\#Break-BERT}}\\
\hline
{Category} & {Precision} & {Recall} & {Precision} & {Recall}\\
\hline
{Poor} & {4.3\%} & {\textbf{28.6\%}} & {\textbf{50.0\%}} & {14.3\%}\\
{Fair} & {26.7\%} & {46.3\%} & {\textbf{49.2\%}} & {\textbf{47.8\%}}\\
{Great} & {87.3\%} & {57.5\%} & {\textbf{89.7\%}} & {\textbf{92.4\%}}\\
\bottomrule
\end{tabular}
\caption{Performance analysis on different categories. '\#' stands for 'Fine-tune'.}
\label{tab:confusing_matric}
\end{table}
When it shows a similar phrase break pattern to the template, it is highly possible to be a correct phrasing. This explains the high precision, low recall for category 3(Great), as well as the high recall, low precision of category 1(Poor). Figure~\ref{fig:break_case_study} shows examples of how the proposed framework well handled diverse break patterns while Against-TTS failed.



\section{Conclusions}
\label{sec:typestyle}
This work presents a new approach to tackling ESL speech phrase break assessment with pre-trained language models (PLMs). The benefits are two-fold: on the one hand, the introduction of PLMs greatly minimizes the requirements for collecting labeled data; on the other hand, the proposed self-supervised learning can handle multiple possible phrase break patterns of the same text. In the future, leveraging PLMs to solve other prosody assessment tasks, like intonation and stress, is well worth researching.

\bibliographystyle{IEEEbib}
\bibliography{strings,refs}

\end{document}